\documentclass[10pt,twocolumn,letterpaper]{article}

\usepackage{cvpr}              
\usepackage{graphicx}
\usepackage{amsmath}
\usepackage{amssymb}
\usepackage{booktabs}
\usepackage{makecell}  
\usepackage{multirow}
\usepackage{textcomp}
\usepackage{comment}
\usepackage[dvipsnames]{xcolor}

\definecolor{cvprblue}{rgb}{0.21,0.49,0.74}
\usepackage[pagebackref,breaklinks,colorlinks,citecolor=cvprblue]{hyperref}

\def\paperID{7631} 
\def\confName{CVPR}
\def\confYear{2024}

\title{Bridging Restoration and Diagnosis: A Comprehensive Benchmark for Retinal Fundus Enhancement}

\author{Xuanzhao Dong$^{1*}$ \enspace Wenhui Zhu $^{1*}$ \enspace Xiwen Chen$^{2}$ \enspace Hao Wang$^{2}$  \enspace  Xin Li$^{1}$\enspace Yujian Xiong$^{1}$ \\
\enspace Jiajun Cheng$^{1}$ \enspace Zhipeng Wang$^{3}$ \enspace Shao Tang$^{3}$ \enspace Oana Dumitrascu$^{4}$ \enspace Yalin Wang$^{1}$ \\
$^{1}$ Arizona State University,  
$^{2}$ Clemson University, 
$^{3}$ LinkedIn Corporation, \\
$^{4}$ Mayo Clinic
}

\begin{document}

\twocolumn[{%
\renewcommand\twocolumn[1][]{#1}%
\maketitle
\begin{center}
    \centering
    \captionsetup{type=figure}
    \includegraphics[width=1.0\textwidth]{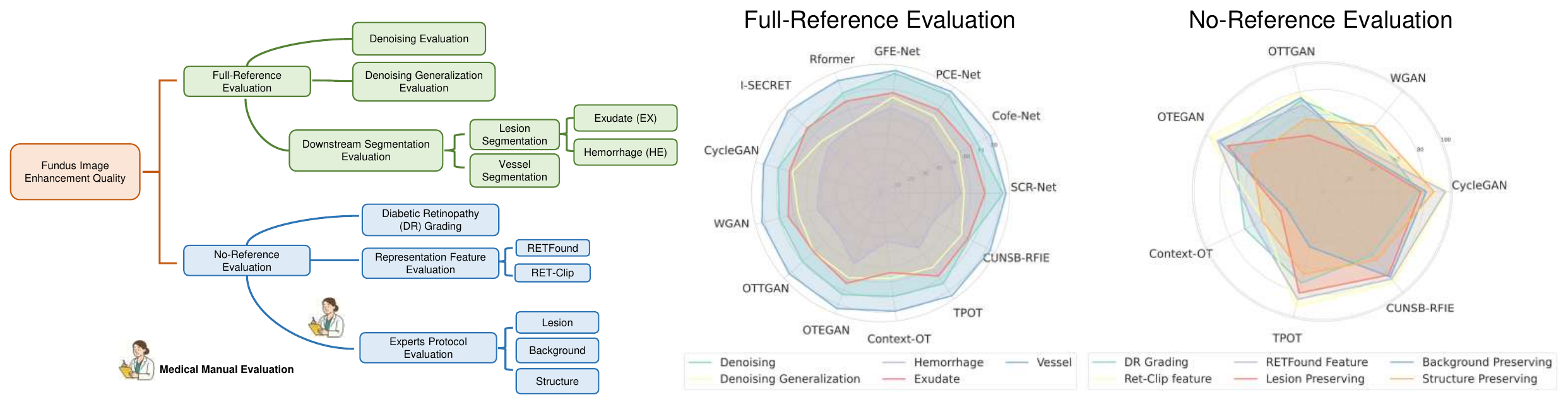}
    \captionof{figure}{ 
    \textbf{Overview of EyeBench-v2.} We present \textbf{EyeBench-v2}, a systematic and comprehensive benchmark designed to evaluate retinal image enhancement models. The evaluation pipeline encompasses both \textbf{Full-Reference} and \textbf{No-Reference} assessments, enabling a robust multi-dimensional analysis of enhancement quality. For each evaluation aspect, we construct a \textbf{distribution-aligned dataset} to ensure fair, reproducible, and clinically relevant comparisons. Additionally, we incorporate \textbf{clinically consistent downstream tasks} to assess models' generalization in denoising and their capacity to preserve diagnostically important features. it also includes expert-guided annotations developed in accordance with established \textbf{clinical protocols}. Our statistical analysis demonstrates that EyeBench-v2 scores strongly align with clinical quality preferences. Finally, EyeBench-v2 facilitates a rigorous and systematic evaluation of existing GAN-based and SDE-based approaches, uncovering \textbf{key limitations} and offering \textbf{actionable insights} into promising solutions for advancing retinal image enhancement.
    }
    \label{banner}
\end{center}%
}]

\def\thefootnote{*}\footnotetext{These authors contributed equally to this paper.}


\begin{abstract}
Over the past decade, generative models have demonstrated remarkable success in enhancing fundus images. However, the evaluation of these models remains a significant challenge. A comprehensive benchmark for fundus image enhancement is critically needed for three main reasons: (1) Conventional denoising metrics such as PSNR and SSIM fail to capture clinically relevant features, such as lesion preservation and vessel morphology consistency, limiting their applicability in real-world settings; (2) There is a lack of unified evaluation protocols that address both paired and unpaired enhancement methods, particularly those guided by clinical expertise; and (3) An ideal evaluation framework should provide actionable insights to guide future advancements in clinically aligned enhancement models. To address these gaps, we introduce \textbf{EyeBench-V2}, a novel benchmark designed to bridge the gap between enhancement model performance and clinical utility. Our work offers three key contributions: \textbf{(1) Multi-dimensional clinical-alignment through downstream evaluations:} Beyond standard enhancement metrics, we assess performance across clinically meaningful tasks including vessel segmentation, diabetic retinopathy (DR) grading, generalization to unseen noise patterns, and lesion segmentation. \textbf{(2) Expert-guided evaluation design:} We curate a novel dataset enabling fair comparisons between paired and unpaired enhancement methods, accompanied by a structured manual assessment protocol by medical experts, which evaluates clinically critical aspects such as lesion structure alterations, background color shifts, and the introduction of artificial structures. \textbf{(3) Actionable insights:} Our benchmark provides a rigorous, task-oriented analysis of existing generative models, equipping clinical researchers with the evidence needed to make informed decisions, while also identifying limitations in current methods to inform the design of next-generation enhancement models.

\end{abstract}

\section{Introduction}
Non-mydriatic retinal color fundus photography (CFP) has become a standard imaging modality in ophthalmology, widely adopted for the analysis of various retinal diseases due to its convenience and the advantage of not requiring pupillary dilation~\cite{deeplearning1,deeplearning2,deeplearning3,deeplearning4,deeplearning5,deeplearning6}. However, the quality of CFP images is often compromised by a range of factors, including imaging artifacts, uneven illumination, ocular media opacities, defocus, or suboptimal acquisition conditions~\cite{shen2020modeling,fu2019evaluation}. In response, the field has seen rapid progress in fundus image enhancement, particularly driven by the advancement of generative models. These models, especially unpaired image-to-image translation approaches, have gained traction due to their ability to learn from unpaired datasets, alleviating the reliance on hard-to-collect paired noisy and clean images~\cite{wang2022optimal,zhu2023optimal,zhu2023otre,dong2024cunsb,vasa2024context}. These unpaired methods have shown competitive performance, and have become increasingly favored by clinicians and researchers due to the practical challenges of collecting perfectly aligned image pairs in clinical settings.

Despite these advancements, the evaluation of fundus image enhancement remains inadequate. Most existing evaluation protocols are inherited from supervised denoising tasks, which typically involve synthesizing noisy-clean image pairs by injecting artificial noise (e.g., Gaussian blur, additive white noise) into high-quality images. These settings rely heavily on conventional metrics (e.g., PSNR and SSIM), which often fail to reflect clinical relevance or capture meaningful differences in high-level representations between enhanced and real high-quality images. Furthermore, enhancement performance alone is insufficient for clinical translation. In a word, a more comprehensive and clinically meaningful evaluation framework is needed, one that rigorously assesses both paired and unpaired enhancement methods. To bridge the gap between image quality enhancement and clinical diagnostic requirements, we present \textbf{EyeBench-V2}, a comprehensive benchmark designed to evaluate fundus image enhancement models from both algorithmic and clinical perspectives. \textbf{EyeBench-V2} offers three key contributions:

First, it introduces a suite of downstream tasks that reflect clinically relevant assessment criteria, thereby decomposing enhancement quality into dimensions aligned with medical preferences. These tasks emphasize the preservation of retinal vessel structures, disease severity grading, and lesion integrity. We implement a unified evaluation pipeline in which existing enhancement methods are trained within a standardized framework and applied to improve fundus image quality prior to task-specific evaluation. As illustrated in Fig.~\ref{banner}, our downstream tasks include enhancement generalization, vessel segmentation, lesion segmentation, image representation evaluation, and diabetic retinopathy (DR) grading. Each task measures the divergence between predicted outputs (e.g., masks, labels, or latent representations) from enhanced images and high-quality images. This enables a precise assessment of whether enhancement methods preserve critical anatomical and pathological features. The inclusion of these clinically aligned tasks not only evaluates enhancement fidelity but also establishes a foundation for assessing the translational potential of enhancement models in real-world diagnostic workflows.

Second, to enable rigorous and fair comparisons, we curate a new dataset with expert annotations of unusable images and resampled disease severity labels for each subset. This dataset supports both full-reference (synthetic) and no-reference (real-world) evaluation scenarios. For full-reference settings, we provide dedicated training and testing splits for both paired and unpaired enhancement methods, facilitating comparative studies across denoising performance and downstream task accuracy. In no-reference scenarios, we restructure the data splits to reflect typical unpaired enhancement use cases in clinical practice. Moreover, we design a medical expert–guided manual evaluation protocol (Fig.~\ref{banner}) that quantitatively assesses clinical quality aspects such as lesion distortion, background shifts, and artificial structure generation. Statistical analysis of expert annotations further validates the importance and reliability of this multi-dimensional evaluation scheme.

Third, the comprehensive results from \textbf{EyeBench-V2} (see Fig.~\ref{banner}) offer actionable insights for medical professionals and researchers, helping to identify the most appropriate enhancement methods to support reliable diagnosis and image interpretation. In particular, we highlight the performance and generalization capacity of clinically relevant unpaired models in denoising and structure preservation. Additionally, \textbf{EyeBench-V2} offers a detailed analysis of current methodological limitations and provides guidance for future research aimed at advancing the clinical utility of fundus image enhancement techniques.


\section{Existing Methods} \label{Sec: methods}
We aim to investigate the effectiveness of current image denoising approaches, with a particular emphasis on both paired and unpaired training paradigms. Let $\mathbf{X}_i$ and $\mathbf{Y}_i$ denote the distributions of low-quality and high-quality fundus images, respectively, where $i \in {1, 2}$ refers to disjoint datasets. For paired methods (described in Sec.~\ref{Sec:supervised}), we consider image pairs $(\mathbf{x}_1, \mathbf{y}_1)$ such that $\mathbf{x}_1 \sim \mathbb{P}_{\mathbf{X}_1}$ and $\mathbf{y}_1 \sim \mathbb{P}_{\mathbf{Y}_1}$, ensuring a direct correspondence. In contrast, for unpaired methods (outlined in Sec.~\ref{Sec:unsupervised}), the training data consists of independent samples $\mathbf{x}_1 \sim \mathbb{P}_{\mathbf{X}_1}$ and $\mathbf{y}_2 \sim \mathbb{P}_{\mathbf{Y}_2}$, with no explicit alignment between the two domains. 

\subsection{Paired Methods}\label{Sec:supervised}
Paired methods for retinal fundus image enhancement can be uniformly formulated as:
\begin{equation}\label{eq:paired-methods}
    \hat{\mathbf{y}}_1 = f_\theta(\mathbf{x}_1)
\end{equation}
\noindent Here, $(\mathbf{x}_1, \mathbf{y}_1)$ denote the paired data, and $f_\theta$ represents the denoising network. These methods typically simulate degradation using predefined noise models and employ various neural architectures to restore image quality. Several representative approaches have incorporated specialized priors into this framework. For example, SCR-Net\cite{li2022structure}, Cofe-Net\cite{shen2020modeling}, PCE-Net\cite{10.1007/978-3-031-16434-7_49}, and GFE-Net\cite{li2023generic} adopt variational autoencoder (VAE)-based frameworks, where $\hat{\mathbf{y}}_1$ is estimated using additional regularization such as high-frequency components, retinal structural priors, artifact maps, or Laplacian pyramid features. 

In contrast, RFormer~\cite{deng2022rformer} introduces a transformer-based architecture for $f_\theta$, emphasizing the modeling of long-range dependencies within the input $\mathbf{x}_1$. Notably, I-SECRET~\cite{i-secret} adopts a semi-supervised training strategy: in the initial stages, paired supervision is used to enforce structural fidelity and pixel-level alignment, followed by unpaired adversarial training, where $f_\theta$ serves as a generative model. Despite its hybrid nature, we categorize I-SECRET as a paired method for consistency in our evaluation framework.

\subsection{Unpaied Methods} \label{Sec:unsupervised}
Unpaired methods for retinal image denoising can be broadly categorized into two primary approaches: GAN-based (e.g.,Generative Adversarial Networks (GANs)\cite{goodfellow2020generative}) and SDE-based models (e.g., Diffusion Models~\cite{ho2020denoising,song2020denoising}, and Gradient Flow-based models~\cite{song2019generative}). Given the practical challenges of acquiring paired clean and degraded retinal images, many unpaired methods reframe denoising as a style transfer problem between the distributions of low- and high-quality images.

\noindent\textbf{GAN-based model}. GAN-based models utilize adversarial learning to generate high-fidelity retinal images that preserve fine anatomical structures. A standard adversarial training objective is expressed as:
\begin{equation}\label{eq:gan-typical}
\begin{split}
    \min_{G_{\mathbf{X}_1}} \max_{D_{\mathbf{Y}_2}} \mathcal{L}  &:= \mathbb{E}_{\mathbf{y}_2}[\log D_{\mathbf{Y}_2}(\mathbf{y}_2)] \\
   & + \mathbb{E}_{\mathbf{x}_1 }[\log(1 - D_{\mathbf{Y}_2}(G_{\mathbf{X}_1}(\mathbf{x}_1)))]
\end{split}
\end{equation}
\noindent Here, $G_{\mathbf{X}_1}$ and $D_{\mathbf{Y}_2}$ represent the generator and discriminator, respectively. 

\noindent\textit{CycleGAN}~\cite{cyclegan} eliminates the need for paired training data by introducing a dual-generator architecture with cycle consistency and identity losses. This enables bidirectional translation and better semantic alignment between domains. However, the added architectural complexity increases computational overhead and may induce failure modes (e.g., mode collapse and structural artifacts), especially when handling images with multimodal content\cite{salmona2022can}.

\noindent\textit{Wasserstein-GAN} (WGAN)~\cite{arjovsky2017wasserstein,gulrajani2017improved} promotes training stability by employing the Wasserstein distance as part of the optimization objective. Rather than solving the primal optimal transport (OT) problem directly, WGAN utilizes the Kantorovich-Rubinstein duality\cite{villani2009optimal} to approximate distribution distance, leading to the following objective:
\begin{equation}\label{eq:wgan-basic}
    \begin{split}
        \min_{G_{\mathbf{X}_1}} \max_{D_{\mathbf{Y}_2}} \mathcal{L} &:= \mathbb{E}_{\mathbf{y}_2 }[D_{\mathbf{Y}_2}(\mathbf{y}_2)] 
        - \mathbb{E}_{\mathbf{x}_1 } [D_{\mathbf{Y}_2}(G_{\mathbf{X}_1}(\mathbf{x}_1))]
    \end{split}
\end{equation}
\noindent Here, the discriminator $D_{\mathbf{Y}_2}$ assigns continuous scores to both real and generated samples, guiding the generator $G_{\mathbf{X}_1}$ to minimize the Wasserstein distance and encouraging smoother, more stable training.

In contrast to WGAN’s dual formulation, \textit{OTT-GAN}~\cite{wang2022optimal} directly addresses the Monge OT problem via an adversarial strategy. The objective function is given by:
\begin{equation} \label{eq:ott-basic}
    \begin{split}
        \max_{G_{\mathbf{X}_1}} \min_{D_{\mathbf{Y}_2}} \mathcal{L}:= \mathbb{E}_{\mathbf{x}_1} [ C( \mathbf{x}_1, G_\theta (\mathbf{x}_1) )] 
        +\lambda \mathbf{W}_1(\mathbb{P}_{\mathbf{Y}_2}, \mathbb{P}_{G_{\mathbf{X}_1}(\mathbf{x}_1)} )
    \end{split}
\end{equation}
\noindent Here, the cost function $C$ is typically defined as mean squared error (MSE), and the Wasserstein distance $\mathcal{W}_1$ is approximated using the WGAN objective. Building on \textit{OTT-GAN}, \textit{OTE} , and \textit{OTRE}~\cite{zhu2023optimal,zhu2023otre} incorporates the Multi-Scale Structural Similarity Index Measure (MS-SSIM)~\cite{wang2003multiscale,brunet2011mathematical} as a perceptual loss to improve the structural consistency of the translated images, alongside identity regularization to better preserve image content. To further improve the perceptual alignment between input and output, \textit{Context-aware OT}~\cite{vasa2024context} extends beyond pixel-based costs by leveraging a pretrained VGG~\cite{mechrez2018contextual} network to compute contextual losses in the feature space, approximating the Earth Mover's Distance over feature representations. Finally, \textit{TPOT}~\cite{dong2024tpot} introduces a topology-aware regularization framework that preserves vascular structures by minimizing the discrepancy between the persistent homology (i.e., topological summaries) of $\mathbf{x}_1$ and $G_{\mathbf{X}_1}(\mathbf{x}_1)$. This approach extends the idea of topological preservation beyond segmentation to directly guide image enhancement.

\noindent\textbf{SDE-based model}. \textit{CUNSB-RFIE}~\cite{dong2024cunsb} models the image enhancement process as the Schr\"{o}dinger Bridge (SB) problem. By simulating the Schr\"{o}dinger Bridge Coupling (SBC) between arbitrary sub-intervals, this approach enables a smooth and probabilistically consistent transformation. However, it can suffer from the progressive attenuation of high-frequency details during iterative training. The main objective function for an arbitrary step $t_i$ is expressed as:
\begin{equation}\label{eq:CUNSB-RFIE}
\begin{split}
    &\min_{\phi} \mathbb{L}(\phi,t_i) := \mathbb{L}_{Adv}(\phi,t_i) + \lambda_{SB} \mathbb{L}_{SB}(\phi,t_i) 
\end{split}
\end{equation}
\noindent Here, $\phi$ denotes the parameter of the generator $G_{\mathbf{X}_1}$. The term $\mathbb{L}_{Adv}$ modifies the KL-divergence between the synthetic high-quality image distribution and the ground-truth distribution $\mathbb{P}_{\mathbf{Y}_2}$. $ \mathbb{L}_{SB}$ serves as an approximation of the entropy-regularized optimal transport, guiding the generator toward a solution that aligns with the SBC.


\section{Clinic Experts Guided Data Annotation}
Color fundus images from the EyeQ dataset~\cite{fu2019evaluation} were utilized throughout image denoising. However, two major issues necessitated re-annotation and resampling before adoption.

First, there is a noticeable \textbf{distribution misalignment} across both image quality categories and diabetic retinopathy (DR) severity grades. This misalignment risks underestimating the true capability of generative models in denoising and lesion preservation. For example, if the test set predominantly consists of "usable" images while the training set includes mostly "good quality" or low-DR images, the models may gain an unfair advantage or fail to generalize properly. Second, we observed a substantial number of \textbf{overprocessed images} within the "usable" and "reject" categories. Training with these collapsed or heavily distorted images may degrade model performance, particularly in learning to remove either synthetic or real-world noise, thereby compromising diagnostic utility. 

To address these issues, we applied comprehensive filtering, resampling, and post-processing procedures under the supervision of medical experts. As a result, we curated two evaluation datasets: the \textbf{Full-Reference Evaluation Dataset}, comprising 16,817 images, is designed to assess model performance under synthetic noise conditions for all algorithms; the \textbf{No-Reference Evaluation Dataset}, consisting of 6,434 subjects, is used to evaluate the unpaired model’s clinical applicability and its alignment with expert preferences. We provide additional details (e.g., dataset specifications, resampling procedures, etc.) in \textcolor{red}{Appendix A}.

\begin{table*}[!t]
\centering
\caption{Performance comparison of denoising evaluation in Full-Reference quality assessment experiments. The best performance in each column is highlighted in bold, with the second-best underlined. Visualization results refer to the \textcolor{red}{Appendix D}.}
\tiny
\resizebox{0.9\textwidth}{!}{%
\begin{tabular}{cccccccc}
\toprule
\multirow{2}{*}{} & \multirow{2}{*}{\textbf{Method}} & \multicolumn{2}{c}{\textbf{EyeQ}} & \multicolumn{2}{c}{\textbf{IDRID}} & \multicolumn{2}{c}{\textbf{DRIVE}} \\ \cmidrule(l){3-8} 
                                          &                                  & \textbf{SSIM} $\uparrow$   & \textbf{PSNR} $\uparrow$   & \textbf{SSIM} $\uparrow$   & \textbf{PSNR} $\uparrow$   & \textbf{SSIM} $\uparrow$   & \textbf{PSNR} $\uparrow$   \\ \midrule
\multirow{5}{*}{\textit{Paired Methods}} & SCR-Net~\cite{li2022structure}   & \textbf{0.9606} & 29.698 & 0.6425 & 18.920 & \textbf{0.6824} & 23.280 \\      
                                         & Cofe-Net~\cite{shen2020modeling} & 0.9408           & 24.907           & 0.7397            & 20.058            & 0.6671            & 21.774            \\
                                         & PCE-Net~\cite{10.1007/978-3-031-16434-7_49} & 0.9487           & \textbf{29.895}           & \underline{0.7764}            & \underline{23.201}           & 0.6704            & 24.041           \\
                                         & GFE-Net~\cite{li2023generic}     & \underline{0.9554}           & \underline{29.719}           & \textbf{0.7935}            & \textbf{25.012}           & \underline{0.6793}            & \underline{23.786} \\
                                         &RFormer~\cite{deng2022rformer}
                                          & 0.9260   & 27.163   & 0.5963   & 18.433   & 0.6311   & 22.172\\
                                          & I-SECRET~\cite{i-secret}        & 0.9051 & 23.483 & 0.7157 & 20.173 & 0.5727 & 18.803 \\
                                          \midrule
\multirow{7}{*}{\textit{Unpaired Methods}} 
                                         & CycleGAN~\cite{cyclegan}         & 0.9313         &\underline{25.076}           & \textbf{0.7668}            & \underline{22.511}           & \underline{0.6681}            & \textbf{22.686}  \\
                                         & WGAN~\cite{gulrajani2017improved} & 0.9266          & 24.793           & 0.7316           & 21.325           & 0.6431            & 20.408 \\
                                         & OTTGAN~\cite{wang2022optimal}    & 0.9275           & 24.065           & 0.7509           & 22.131           & 0.6635            & 21.938 \\
                                         & OTEGAN~\cite{zhu2023optimal}     & \underline{0.9392}           & 24.812         & 0.7624           & 22.272           & 0.6642            & 22.183 \\
                                         & Context-aware OT~\cite{vasa2024context} & 0.9144           & 24.088           & 0.7338            & 21.790            & 0.6407            & 21.389 \\
                                         & TPOT~\cite{dong2024tpot} & \textbf{0.9417} & \textbf{25.196} & 0.7636 & \textbf{22.556} & \textbf{0.6731} & 22.142 \\
                                         & CUNSB-RFIE~\cite{dong2024cunsb}  & 0.9121           & 24.242           & \underline{0.7651}           & 22.448           & 0.6659            & \underline{22.510} \\
\bottomrule
\end{tabular}%
}
\label{tb:deg-exp}
\vspace{-0.1cm}
\end{table*}

\begin{table*}[!t]
    \centering
    \caption{
    Performance comparison of vessel and lesion (EX and HE) segmentation in Full-Reference quality assessment experiments. The best performance in each column is highlighted in bold, with the second-best underlined. For visualization results, refer to the \textcolor{red}{Appendix D}.}
    \tiny
    \resizebox{0.9\textwidth}{!}{%
    \begin{tabular}{lcccc|ccc|ccc}
        \toprule
         \multirow{2}[3]{*}{Method} & \multicolumn{4}{c}{Vessel Segmentation} & \multicolumn{3}{c}{EX} & \multicolumn{3}{c}{HE} \\ 
         \cmidrule(lr){2-5}  \cmidrule(lr){6-8}  \cmidrule(lr){9-11}
          
         & AUC $\uparrow$ & PR $\uparrow$ & F1 Score $\uparrow$ & SP $\uparrow$ & AUC  & PR & F1 Score  & AUC & PR & F1 Score \\ \midrule

        SCR-Net~\cite{li2022structure}   & \textbf{0.9227} & \textbf{0.7783} & \textbf{0.7000} & 0.9787 & \textbf{0.9683} & \textbf{0.6041} & \textbf{0.5556} & 0.9377 & 0.3213 & 0.3725\\ 
        cofe-Net~\cite{shen2020modeling} & \underline{0.9188} & \underline{0.7698} & 0.6895 & 0.9801 & 0.9623 & 0.5620 & 0.5349 & 0.9302 & 0.3152 & 0.3281\\
        PCE-Net~\cite{10.1007/978-3-031-16434-7_49} & 0.9146 & 0.7616 & 0.6790 & \textbf{0.9814} & \underline{0.9667} & \underline{0.5876} & 0.5066 & \underline{0.9545} & \underline{0.3639} & \underline{0.3736}\\
        GFE-Net~\cite{li2023generic} & 0.9175 & 0.7669 & 0.6832 & \textbf{0.9814} & 0.9560 & 0.5548 & \underline{0.5380} & \textbf{0.9577} & \textbf{0.4113} & \textbf{0.3751}\\
        RFormer~\cite{deng2022rformer} & 0.8990 & 0.7239 & 0.6374 & \underline{0.9806} & 0.9626 & 0.5593 & 0.4692 & 0.9207 & 0.2677 & 0.3136 \\
        I-SECRET~\cite{i-secret} & 0.9181 & 0.7662 & 0.6838 & 0.9802 & 0.9613 & 0.5424 & 0.4825 & 0.9028 & 0.2629 & 0.2642 \\
    \midrule
        CycleGAN~\cite{cyclegan} & 0.9015 & 0.7278 & 0.6462 & 0.9801  & 0.9447& 0.4843 & 0.4790 & 0.8970 & 0.1624 & 0.2227\\
        WGAN~\cite{gulrajani2017improved} & 0.9081 & 0.7494 & 0.6768 & 0.9764  & 0.9522 & 0.4942 & 0.4859 & \underline{0.8990} & 0.1847 & \underline{0.2476}\\
        OTTGAN~\cite{wang2022optimal} & 0.9034 & 0.7400 & 0.6609 & \underline{0.9812}  & 0.9492 & 0.4214 & 0.4365 & 0.8179 & 0.1448 & 0.2233 \\
        OTEGAN~\cite{zhu2023optimal} & 0.9156 & \underline{0.7678} & \underline{0.6919} & 0.9797 & 0.9562 & 0.5191 & 0.4868 & \textbf{0.9359} & \textbf{0.2800}& \textbf{0.3165} \\
        Context-aware OT~\cite{vasa2024context} & 0.8871 & 0.7077 & 0.6377 & 0.9791 & 0.9305 & 0.3318 & 0.3707 & 0.8091 & 0.0646 & 0.1184 \\
        TPOT~\cite{dong2024tpot} & \textbf{0.9191} & \textbf{0.7748} & \textbf{0.6926} & \textbf{0.9816} & \textbf{0.9615} & \textbf{0.5487} & \textbf{0.5238} & 0.8927 & \underline{0.2110} & 0.2446 \\
        CUNSB-RFIE~\cite{dong2024cunsb}  & \underline{0.9163} & 0.7626 & 0.6872 & 0.9784 & \underline{0.9572}& \underline{0.5381} & \underline{0.4883} & 0.8488 & 0.1489 & 0.1893 \\
        \bottomrule
    \end{tabular}}
    \label{tab-seg}
    \vspace{-0.3cm}
\end{table*}

\begin{table*}[!t]
    \centering
    \caption{Performance comparison with unpaired baselines in No-Reference quality assessment task. The best performance in each column is highlighted in bold, and the second-best is underlined. Visualization results refer to \textcolor{red}{Appendix D}.}
    \tiny
    \resizebox{0.9\textwidth}{!}{%
    \begin{tabular}{lcccc|cc|ccc}
        \toprule
         \multirow{2}[3]{*}{Method} & \multicolumn{4}{c}{DR grading} & \multicolumn{2}{c}{Representation Feature} & \multicolumn{3}{c}{Experts Protocol Evaluation} \\ 
         \cmidrule(lr){2-5}  \cmidrule(lr){6-7}  \cmidrule(lr){8-10}
          
         & ACC $\uparrow$ & Kappa $\uparrow$ & F1 Score $\uparrow$ & AUC $\uparrow$ & FID-Retfound~\cite{zhou2023foundation}$\downarrow$  & FID-Clip~\cite{du2024ret} $\downarrow$ & LPR $\uparrow$ & BPR $\uparrow$ & SPR $\uparrow$\\ \midrule
         
        CycleGAN~\cite{cyclegan}                & \textbf{0.7588} & \underline{0.6006} & \underline{0.7180} & \underline{0.9251}  & \textbf{23.778}& \underline{11.530}  & 0.7707 & 0.8153 & \textbf{0.8726}\\
        
        WGAN~\cite{gulrajani2017improved}       & 0.6446 & 0.3123 & 0.6156 & 0.8874  & 74.885 & 33.076  & 0.4076 & 0.4204 & 0.6561\\
        
        OTTGAN~\cite{wang2022optimal}           & 0.7440 & 0.5688 & 0.7037 & 0.9247  & 51.201 & 20.505 & 0.4586 & 0.7580 & 0.5860 \\
        
        OTEGAN~\cite{zhu2023optimal}            & \underline{0.7539} & \textbf{0.6433} & \textbf{0.7228} & \textbf{0.9326} & \underline{28.987} & \textbf{11.114} & \textbf{0.8280} & \textbf{0.8981} & 0.6178 \\
        
        Context-aware OT~\cite{vasa2024context} & 0.7301 & 0.3811 & 0.6662 & 0.9112 & 61.429 & 34.456 & 0.3566 & 0.3121 & 0.5159 \\

        TPOT~\cite{dong2024tpot} & 0.7169 & 0.5717 & 0.6912 & 0.9232 &34.331 &13.780&\underline{0.8089}&0.4395&0.6624 \\
        
        CUNSB-RFIE~\cite{dong2024cunsb}         & 0.6565 & 0.3674 & 0.6341 & 0.8927 & 33.047 & 14.827  & \textbf{0.8280} & \underline{0.8535} & \underline{0.6879} \\

        \bottomrule
    \end{tabular}}
    \label{tab-noref}
    \vspace{-0.3cm}
\end{table*}

\begin{figure*}[t] \centering \includegraphics[width=0.9\textwidth]{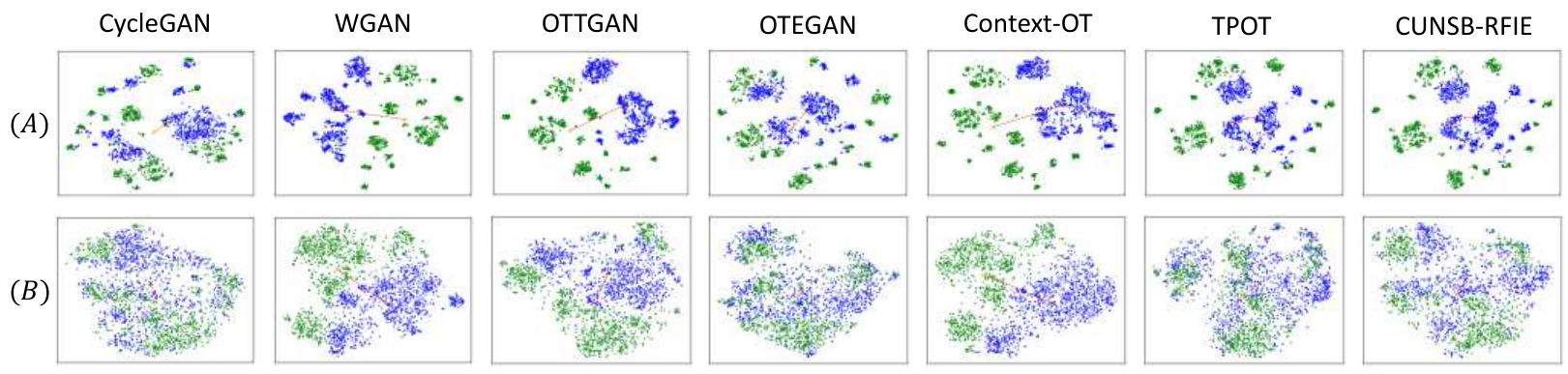}  
\caption{T-SNE~\cite{van2008visualizing} visualizations of the latent representation features extracted from the RETfound \textbf{(A)}~\cite{zhou2023foundation} and Ret-Clip \textbf{(B)}~\cite{du2024ret} image encoder in no-reference evaluation. Here, blue points illustrate synthetic high-quality image $\hat{\mathbf{y}}_1$ features while green points show true high-quality image $\mathbf{y}_2$ features. Closer proximity of the distributions indicates improved denoising performance of the unpaired method. More details provided in Sec.~\ref{subsec:exp}.}
\label{fig:tsne} 
\vspace{-0.3cm}
\end{figure*}

\section{Experiments}
\subsection{Full-Reference Quality Assessment Experiments} \label{section-4.1}
For full-reference evaluation, we used the Full-Reference Evaluation Dataset and adhered strictly to the original training configurations for both paired and unpaired methods. In the unpaired setting, synthetic low-quality images were used as inputs, with unpaired high-quality images as targets. For the paired setting, models were trained in a supervised manner. All models were trained using hyperparameters as reported in their respective papers. For the segmentation tasks, a vanilla U-Net~\cite{ronneberger2015unet} was trained from scratch.

The evaluation included the following baselines: \textit{Paired method}: SCR-Net~\cite{li2022structure}, Cofe-Net~\cite{shen2020modeling}, PCE-Net~\cite{10.1007/978-3-031-16434-7_49}, GFE-Net~\cite{li2023generic}, and RFormer~\cite{deng2022rformer}; \textit{Unpaired method}: I-SECRET~\cite{i-secret}, CycleGAN~\cite{cyclegan}, WGAN~\cite{gulrajani2017improved}, OTTGAN~\cite{wang2022optimal}, OTEGAN~\cite{zhu2023optimal}, Context-aware OT~\cite{vasa2024context}, TPOT~\cite{dong2024tpot}, and CUNSB-RFIE~\cite{dong2024cunsb}. Enhanced images generated from each model were used for downstream evaluations. See \textcolor{red}{Appendix B} for implementation details.

\noindent \textbf{Denoising Evaluation.} Noisy images from the Full-Reference testing set were processed to generate enhanced outputs, which were assessed using Peak Signal-to-Noise Ratio (PSNR) and Structural Similarity Index Measure (SSIM).

\noindent \textbf{Denoising generalization Evaluation.} To assess generalization ability, high-quality images from DRIVE~\cite{drive} and IDRID~\cite{idrid} were synthetically degraded, and the resulting low-quality images were denoised using the trained models in denoising evaluation. PSNR and SSIM were computed between the enhanced and original high-quality images.

\noindent \textbf{Vessel Segmentation}. Using the DRIVE dataset with ground-truth vessel masks, we evaluated structural preservation by training and testing a segmentation model on enhanced images from the denoising generalization task. The dataset was split into 20 training and 20 testing subjects. Metrics include the Area Under the ROC Curve (AUC), the Area under the Precision-Recall Curve (PR), F1 Score, and Specificity (SP).

\noindent \textbf{Lesion Segmentation}. For lesion segmentation, we used the IDRID dataset with annotated masks. Only prominent lesion types (i.e., Hard Exudates (EX) and Hemorrhages (HE) ) were considered. The model was trained and tested on enhanced images using 54 training and 27 testing subjects, respectively. Evaluation metrics include AUC, PR, and F1 score.

\subsection{No-Reference Quality Assessment Experiments}
Evaluating enhancement without ground-truth clean images is challenging for paired methods, thus, we focused on unpaired approaches to assess their real-world denoising performance. The enhanced outputs were evaluated through downstream tasks, including DR grading, feature representation analysis, and expert review. Baselines included CycleGAN~\cite{cyclegan}, WGAN~\cite{gulrajani2017improved}, OTTGAN~\cite{wang2022optimal}, OTEGAN~\cite{zhu2023optimal}, Context-aware OT~\cite{vasa2024context}, TPOT~\cite{dong2024tpot}, and CUNSB-RFIE~\cite{dong2024cunsb}. All models were trained from scratch using their original configurations and training details are outlined in \textcolor{red}{Appendix C}.

\noindent\textbf{DR grading.} An NN-MobileNet~\cite{deeplearning1} was trained on high-quality images for DR classification. Enhanced test images were used for inference, and performance was assessed using accuracy (ACC), kappa score, F1 score, and AUC. This task evaluates whether denoising alters lesion features in ways that affect DR grade consistency.

\noindent\textbf{Representation Feature Evaluation.} We evaluated the similarity of enhanced and high-quality images using Fréchet Inception Distance (FID) computed on two fundus foundation models: Retfound~\cite{zhou2023foundation} and Ret-Clip~\cite{du2024ret}, denoted as \textit{FID-Retfound} and \textit{FID-Clip}. \textit{FID-Retfound}, based on a MAE backbone, captures high-level semantic structures, while \textit{FID-Clip}, trained via contrastive learning, emphasizes spatial coherence and structural consistency.


\noindent\textbf{Experts Annotation Evaluation.} To align with clinical preferences, we conducted expert-guided evaluations using three metrics: Background Preserving Ratio (BPR), Lesion Preserving Ratio (LPR), and Structure Preserving Ratio (SPR), which quantify changes in background, lesion regions, and structural features, respectively. Rather than using the entire test set, we selected 157 images with prominent lesions (DR grades 2-4) to focus on clinically relevant cases. This evaluation task serves to assess the real-world applicability of unpaired denoising models. Detailed protocols are provided in \textcolor{red}{Appendix C}.

\begin{figure*}[t] \centering \includegraphics[width=\textwidth]{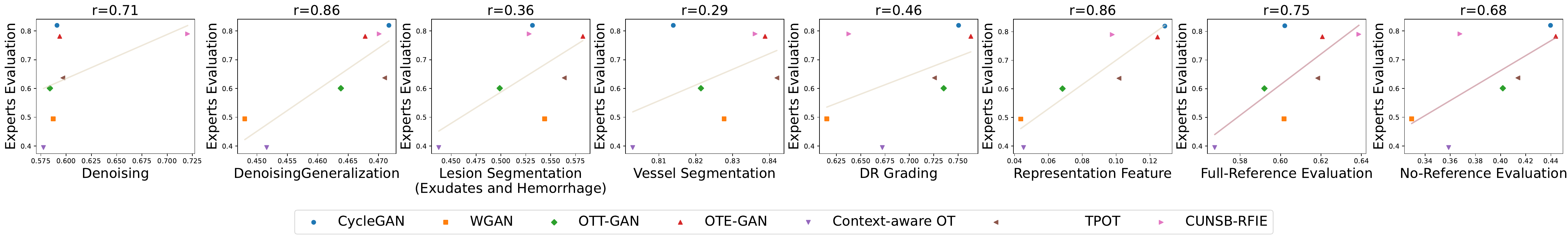}  
\caption{Validation of Expert Clinic Preference Alignment via Spearman’s correlation coefficient ($r$) between the medical experts (i.e., Experts Protocol Evaluation) and other tasks. Notably, our multi-dimensional evaluations (e.g., Full-Reference, No-Reference) demonstrated stronger correlation.} 
\label{fig:correlation} 
\vspace{-0.2cm}
\end{figure*}


\subsection{Experiment Results}\label{subsec:exp}

\noindent\textbf{Full-Reference Evaluation.} While paired methods tend to achieve superior performance on standard metrics due to access to strictly aligned supervision, such advantages are of limited relevance in real-world clinical scenarios. As shown in Tab.~\ref{tb:deg-exp}, paired methods such as GFE-Net leverage frequency-domain cues effectively, yielding high performance. However, the requirement for pixel-aligned image pairs makes these methods impractical for many clinical applications, where such data are often unavailable.

In contrast, unpaired methods offer greater practical utility and still achieve competitive performance. Notably, TPOT show leading performance in EyeQ, IDRID nad DRIVE dataset, indicating strong generalization and denoising capability. Furthermore, in downstream segmentation tasks (Tab.~\ref{tab-seg}), TPOT also outperforms alternatives, achieving the higher performance in both vessel and EX lesion segmentation, thus reinforcing its effectiveness in clinically relevant contexts. We further distill three key insights from the behavior of unpaired methods:

\textbf{Smooth distribution transitions promote both contextual structural preservation and domain alignment}. For example, TPOT leverages optimal transport (OT) theory to enable continuous and coherent transformations, while CUNSB-RFIE achieves similar results by solving a relaxed OT formulation. In contrast, although CycleGAN demonstrates strong generalization and denoising, its reliance on strict bidirectional mappings often compromises anatomical fidelity, particularly for fine structures like blood vessels and lesions.

\textbf{Task-specific regularization is critical in the effectiveness of OT-based generative models}. For instance, incorporating topology-aware regularization during fine-tuning enables TPOT to consistently outperform OTEGAN, which shares the same backbone but lacks such regularization. This highlights the importance of integrating clinically informed priors to enhance model applicability in real-world medical scenarios.

\textbf{A trade-off exists between structural preservation and denoising performance}. Overemphasizing structure-preserving objectives (e.g., preserving vessels or lesion structures) may limit the model’s ability to suppress noise, while focusing too heavily on denoising can result in the loss of clinically relevant features. This underscores the need for careful balancing of objective weights and regularization parameters. Additionally, the choice of cost function (i.e., function $C$ in Eq.~\ref{eq:ott-basic}) is equally critical. For example, SSIM-based costs, as used in OTEGAN and TPOT, are more effective in preserving structural integrity than simpler metrics (e.g., MSE), which are used in models like OTTGAN.

\begin{figure*}[t]
     \centering
     \includegraphics[width=1\linewidth]{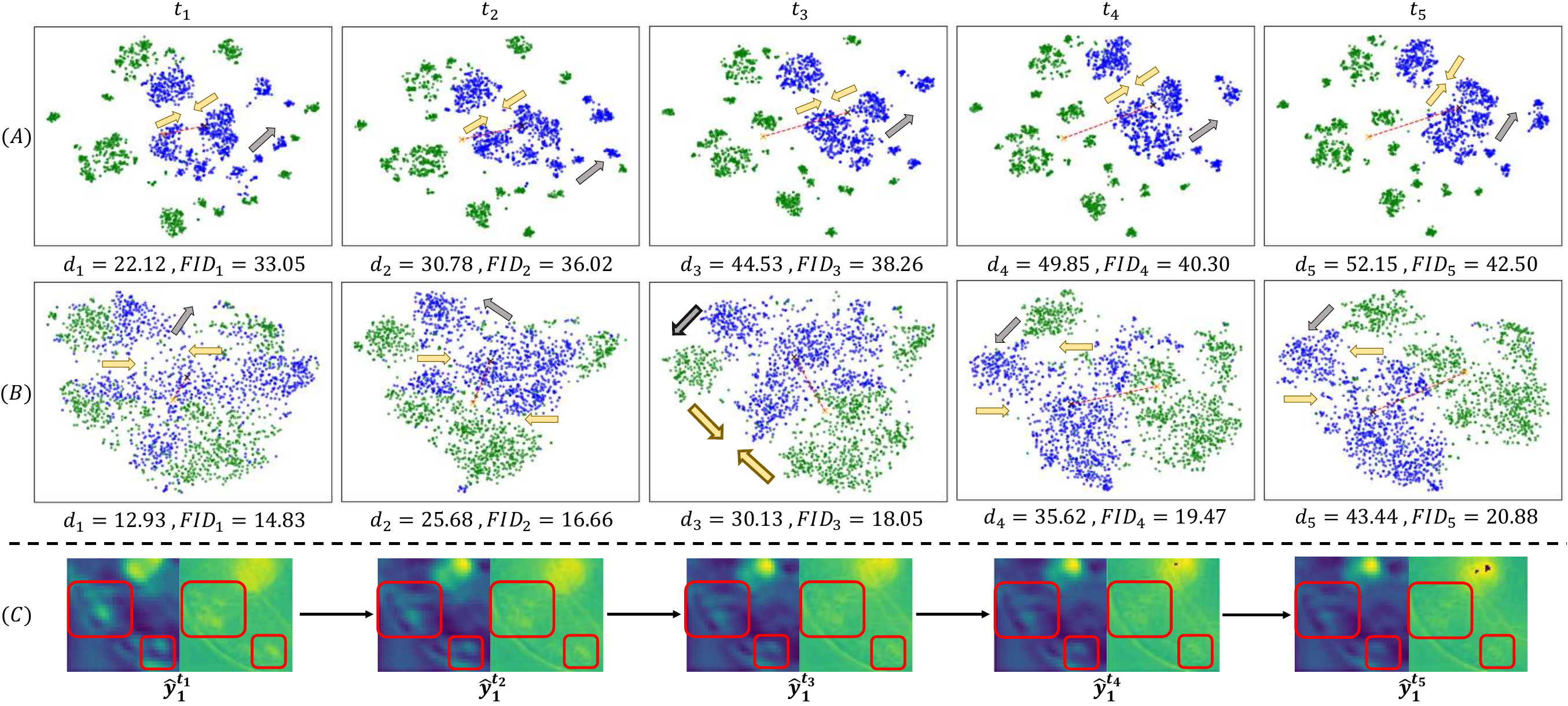}
     \caption{Illustration of the limitations of SDE-based methods. \textbf{(A)} and \textbf{(B)} show T-SNE~\cite{van2008visualizing} visualizations of features from synthetic high-quality images $\hat{\mathbf{y}}_1^{t_i}$ (blue points) and true high-quality images $\mathbf{y}_2$ (green points), extracted using the RETFound~\cite{zhou2023foundation} and Ret-Clip~\cite{du2024ret} image encoders, respectively. Yellow arrows indicate the squeezing effect, while gray arrows denote feature drift over time. $d_i$ and $FID_i$ represent the center distance and corresponding FID score at time step $t_i$. \textbf{(C)} visualizes attention drift in the skip-connection features maps of the generator~\cite{dong2024cunsb} as the time step increases. Red boxes highlight lesion structures that gradually receive less attention. More details are discussed in Sec.~\ref{Sec:further-analysis}.
     }
     \label{fig:SB-noreference}
     \vspace{-0.4cm}
 \end{figure*}

\noindent\textbf{No-Reference Evaluation.} Tab.~\ref{tab-noref} compares the performance of unpaired algorithms under real-world scenarios. It is evident that OTEGAN and CycleGAN consistently achieve strong results across the experiments. For instance, OTEGAN attained the highest kappa, F1 score, and AUC in the DR grading task. Additionally, as shown in T-SNE~\cite{van2008visualizing} visualization (Fig.~\ref{fig:tsne}), they show better feature alignment. These experimental results support the insights previously discussed.

First, bidirectional regularization improves denoising performance and structural integrity by compensating for contextual information loss. For example, CycleGAN, which incorporates cycle consistency loss, demonstrates strong denoising capabilities (reflected in superior DR grading metrics), enhanced feature alignment (lower FID score), and the best overall structural preservation (highest SPR). However, it tends to modify lesion structures more noticeably, resulting in a relatively lower LPR score. Second, optimal cost function selection and smooth distributional transitions, as employed by OTEGAN, contribute to its overall superior performance.

Nevertheless, the findings also raise some concerns. First, the effectiveness of topology-based regularization in real-world scenarios remains uncertain. Although TPOT performs well under synthetic noise conditions, it lags behind other OT-based methods in this real-world setting. This discrepancy underscores the importance of our multi-dimensional evaluations pipeline. Second, while CUNSB-RFIE achieves higher expert preference scores (as reflected in the expert protocol evaluation), it underperforms in the other two tasks. This naturally prompts the question: \textbf{\textit{Why does this SDE-based method fall short compared to GAN-based approaches?}} We delve deeper into this issue in the next section.


\section{Further Analysis}\label{Sec:further-analysis}

\noindent\textbf{The necessity of multi-dimensional evaluation.} 
We want to emphasize the necessity of our multi-dimensional evaluation framework. First, relying on a single evaluation pipeline or metric often provides a limited view and can hinder a comprehensive understanding of generative models in retinal image restoration. For instance, while CycleGAN performs well in denoising and generalization, it fails to adequately preserve contextual and structural information. Similarly, although TPOT achieves near state-of-the-art performance under synthetic noise conditions, it struggles in real-world noise scenarios. Second, evaluation based solely on partial statistical metrics may not reflect true clinical preferences, thereby limiting their practical applicability. As illustrated in Fig.~\ref{fig:correlation}, performance on a single task, such as vessel segmentation, shows weak alignment with clinicians' overall perception of image quality. In contrast, our multi-dimensional evaluation, such as full-reference experiments, demonstrates a stronger and more clinically meaningful correlation.

\noindent\textbf{Limitation in SDE-based method}. SDE-based approaches (e.g., diffusion models) have achieved remarkable success and become standard baselines in natural image generation~\cite{rombach2022high}. However, their application in unpaired retinal fundus image enhancement remains limited. As shown in Tab.~\ref{tb:deg-exp}, \ref{tab-seg} and \ref{tab-noref}, although CUNSB-RFIE demonstrate promising denoising and generalization capabilities, it still underperforms compared to GAN-based approaches, particularly in tasks that require high contextual or semantic alignment. 

To better understand the intrinsic limitations of CUNSB-RFIE, we conducted a series of exploratory experiments. As shown in Fig.~\ref{fig:SB-noreference}(A) and (B), we extracted features from both the synthetic high-quality images $\hat{\mathbf{y}}_1^{t_i}$ and the fixed high-quality images $\mathbf{y}_2$ using the RetFound and Ret-Clip image encoders, and visualized them via T-SNE. Two key observations emerged: First, as the time step $t_i$ increases, both the center distance and the FID score increase, suggesting growing divergence. Second, while features of $\mathbf{y}_2$ (i.e., green dots) remain fixed, the features of $\hat{\mathbf{y}}_1^{t_i}$ exhibit two common transformation patterns: \textbf{Progressive squeezing} into a tighter cluster (yellow arrow), and \textbf{Gradual drift} away from the manifold of $\mathbf{y}_2$ (gray arrow), following either a unified (A) or random (B) direction. These trends motivate the following insights:

First, the observed feature "squeezing" effect reflects the smooth and entropy-optimal learning process intrinsic to SB. Specifically, SB models the most likely evolution of distributions from $\mathbb{P}_{\mathbf{X}_1}$ to $\mathbb{P}_{\mathbf{Y}_2}$ by minimizing the KL-divergence with respect to a Wiener process. To achieve this, CUNSB-RFIE learns the optimal marginal distribution $\pi^*_{(t_i, t_{\text{stop}})}$ as the system evolves from the initial state $t_1$ to a fixed terminal state $t_{\text{stop}}$~\cite{dong2024cunsb,kim2023unpaired}. Consequently, samples draw from $\hat{\mathbf{y}}_1^{t_i} \sim \pi^*_{t_{\text{stop}} | t_i}$ reflect increasingly refined statistics, which evident in the shrinking variance or compressed feature manifold. This squeezing indicates that the model collapses distribution $\pi^*_{t_{\text{stop}} | t_i}$ into lower-entropy region when seeking the most KL-optimal path.

However, a natural question arises: \textbf{\textit{ Is the learned distribution $\pi^*_{t_{\text{stop}} | t_i}$ semantically aligned with the target high-quality manifold?}} Our findings suggest it is not. We argue that this misalignment stems from an inherent trade-off in SB models: \textbf{between smooth probabilistic transitions and semantic fidelity}. Since the Wiener process (i.e., Brownian motion) represents the most random and structure-agnostic prior, the learned transition path favors smooth, low-frequency evolutions over those requiring sharp, semantically meaningful transformations. In medical image enhancement, particularly for tasks that involve preserving high-frequency diagnostic structures such as lesions, this inductive bias may lead to suboptimal outcomes. In contrast, GAN-based models or even the first-step generator in CUNSB-RFIE often perform better in such cases, as they are still driven to model the underlying distribution more directly, rather than defaulting to the "smoothest" path as training progresses.

This interpretation is further supported by empirical evidence. First, as shown in Fig.~\ref{fig:SB-noreference}(C), lesion structures, which typically manifest as high-frequency details in retinal fundus images~\cite{chu2024improving,qiong2025medical} and are often preserved via skip connections in U-Net architectures~\cite{si2024freeu}, were examined via averaged skip connection feature maps from different layers of the generator (as in~\cite{dong2024cunsb}). For images with noticeable structure, we observed that as $t_i$ increases, the model progressively deactivates lesion-relevant regions and shifts its attention away from diagnostically important features, which consistent with path drift. Second, the observed increases in both distributional distance and FID score provide further quantitative evidence of global misalignment between the synthetic and true high-quality distributions. Although these observations arise during inference, it is crucial to note that the CUNSB-RFIE solver employs the same forward generation dynamics during training. As a result, these limitations not only manifest at test time but also fundamentally constrain the model’s capacity to learn an optimal transformation.

We believe that SDE-based methods hold significant promise for the future of retinal image enhancement. However, to fully realize their potential, an important question naturally arises: \textbf{\textit{What are the most promising directions for improvement?}} Based on our findings, we propose the following insights: \textbf{(i)Intergrate additional guidance to encourage diversity exploration. } Incorporating information-rich latent representations from pretrained foundation models may help promote semantically meaningful diversity and improve feature alignment with clinically relevant structures. \textbf{(ii) Explore alternative SDE solvers.} Adopting alternative numerical solvers for SDEs may offer the ability to relax or bypass the limitations imposed by standard Wiener process priors. This could enable the modeling of more expressive or semantically guided transition paths, potentially preserving high-frequency details more effectively. We leave these directions as key avenues for future exploration.

\section{Conclusion}
 
With the rapid advancement of generative models, it has become increasingly important to align image denoising methods for fundus images with practical clinical needs. In this work, we introduce \textit{Eyebench-v2}, a benchmark aimed at providing more rigorous and clinically meaningful evaluations of enhanced images, thereby facilitating broader engagement from medical professionals. Notably, our multi-dimensional evaluation framework exhibits strong agreement with expert manual assessments, underscoring its potential to bridge the gap between generative denoising models and real-world clinical applications. Additionally, our insights derived can guide future research in this domain.



{
    \small
    \bibliographystyle{ieeenat_fullname}
    \bibliography{main}
}

 \clearpage\section*{\Large Supplementary Materials - Bridging Restoration and Diagnosis: A Comprehensive Benchmark for Retinal Fundus Enhancement}
 \begin{figure*}[h]
  \centering
  \includegraphics[width=\textwidth]{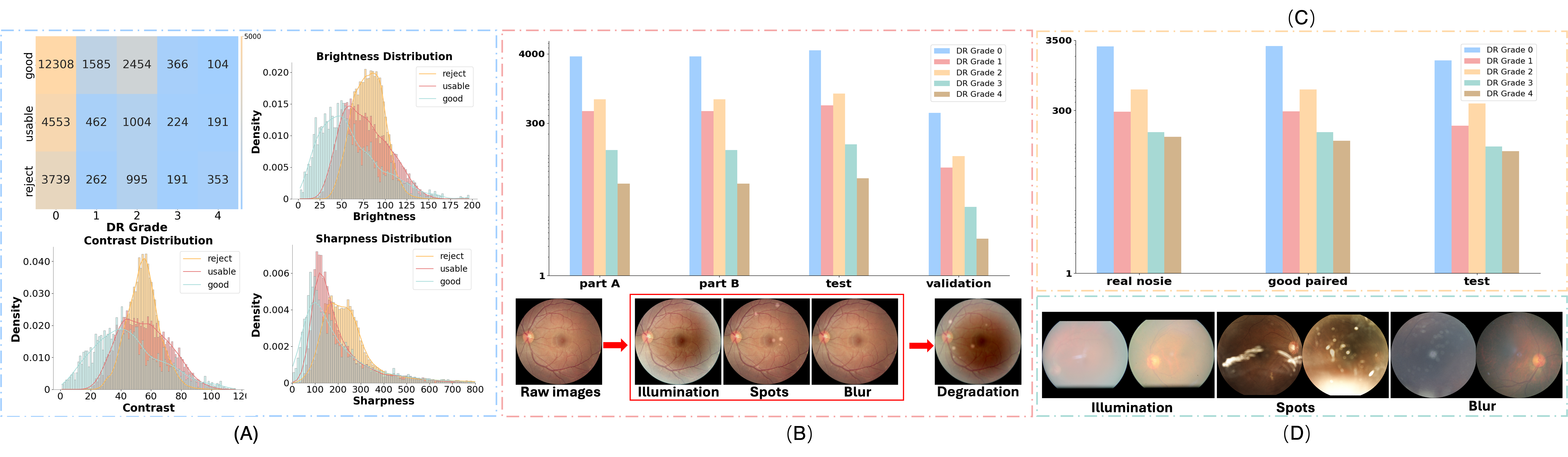}  
  \caption{ Overview of EyeQ~\cite{fu2019evaluation} dataset. (A) highlights attribute distributions (i.e., brightness, contrast, sharpness) and diabetic retinopathy (DR) grades across quality categories (i.e., good, usable, and reject). (B) illustrates histograms for the training (i.e., part A and part B), testing, and validation datasets used in \textbf{Full-Reference} evaluations after resampling, with the workflow of degradation algorithms outlined below. (C) shows histograms for real-world \textbf{No-Reference} experiments after resampling. (D) presents reject-quality samples (e.g., overprocessed images).
  }
  \label{fig:overall_ana}
\end{figure*}

\thispagestyle{empty}
\appendix
\section{Clinical Experts Guided Data Annotation Details}
We utilized 28,791 color fundus images from the EyePACS initiative~\cite{diabetic-retinopathy-detection}, with image quality labels obtained from the EyeQ dataset~\cite{fu2019evaluation}. Each image in the dataset~\cite{fu2019evaluation} was originally assigned a quality category (i.e., good, usable, or reject) and a diabetic retinopathy (DR) severity grade ranging from 0 to 4. As shown in Fig.~\ref{fig:overall_ana}(A), brightness, contrast, and sharpness distributions vary across quality levels, with good and usable images exhibiting similar patterns, while rejected images display distinct characteristics, such as increased sharpness. DR label distribution is imbalanced, with grades 0 and 2 being most frequent and severe cases underrepresented. Overprocessed images were also observed in usable and reject categories, potentially affecting diagnostic utility (Fig.~\ref{fig:overall_ana}(D)). To ensure quality and lesion preservation, we retained only good and usable images, applying ratio-preserving resampling under expert guidance. Due to the scarcity of severe DR cases, we preserved the natural DR label distribution to reflect clinical data characteristics.

\noindent \textbf{Full-Reference Evaluation Dataset.}
We used 16,817 good-quality images, split into 10,000 for training, 600 for validation, and 6,217 for testing (i.e., Fig.~\ref{fig:overall_ana}(B)). All images were synthetically degraded following~\cite{shen2020modeling}, simulating illumination issues, spot artifacts, and blurring. The training set was divided into two disjoint subsets (i.e., $A$ and $B$, each with 5,000 images), and corresponding degraded versions (i.e., $A^{\ast}$ and $B^{\ast}$) were generated. We strictly followed paired  (i.e., $A^{\ast} \rightarrow A$) and unpaired (i.e., $A^{\ast} \rightarrow B$) training pipeline for fair comparison.

\noindent \textbf{No-Reference Evaluation Dataset.}
We selected 6,434 usable-quality images (i.e., Fig.~\ref{fig:overall_ana}(C)), resampling 4,000 for training and 2,434 for testing under real-world noise conditions. Additionally, 4,000 unpaired good-quality images were resampled from the original training pool, with DR label matching to support unpaired training protocols.

\section{Full-Reference Quality Assessment Experiments Details }\label{Sec:full-reference}

For full-reference assessment, we used the previously synthesized Full-Reference Evaluation Dataset. We strictly followed the training configurations for paired and unpaired methods. For the unpaired method, synthetic low-quality images from the training set $A$ (i.e., $A^{\ast}$) were used as input images, while high-quality images from the training set $B$ served as the clean reference images. For the paired method, we performed supervised training using low-high-quality image pairs from the training set $A$ (i.e., $A^{\ast}$ and $A$).

\subsection{SCR-Net~\cite{li2022structure}}
The model was trained for 150 epochs using Adam optimizer, with an initial learning rate of $2 \times 10^{-4}$ and $\beta_1$ value set to $0.5$, followed by 50 epochs with a learning rate linearly decayed to $0$. The training batch size was 32. All images were resized to $ 256 \times 256$ with a random flipping data augmentation technique. For model architectures, the generator and discriminator architectures followed the architectures and configurations described in~\cite{li2022structure}.



\subsection{Cofe-Net~\cite{shen2020modeling}}

The model was trained for 300 epochs using the SGD optimizer, with an initial learning rate of $1 \times 10^{-4}$, which was gradually reduced to 0 over the final 150 epochs. The training batch size was 16, and all images were resize to $512 \times 512$.

The loss function comprised four components: main scale error loss ($L_m$), multiple-scale pixel loss ($L^s_p$), multiple-scale content loss ($L^s_c$) and RSA module loss ($L_v$), as described in~\cite{shen2020modeling}, where the $s$ denotes the scale index. The weight for $L^s_p$, $L^s_c$ and $L_v$ was set to $\lambda_p=10$, $\lambda_c=1$ and $\lambda_v=0.1$, respectively, during the training process.

\subsection{PCE-Net~\cite{10.1007/978-3-031-16434-7_49}}

The model was trained for 200 epochs using the Adam optimizer, with an initial learning rate of $1 \times 10^{-3}$, which was gradually reduced to 0 over the final 50 epochs. The training batch size was 4, and all input images were resized to $256 \times 256$. Data augmentation strategies, including random horizontal and vertical flips with a probability of 0.5, were applied to enhance generalization.

The loss function comprised two components: enhancement loss ($L_E$) and the weighted feature pyramid constraint loss ($L_C$), as described in~\cite{10.1007/978-3-031-16434-7_49}. The weight for $L_C$ was set to $\lambda_C=0.1$ during the training process. Additionally, we adopted a U-Net architecture proposed in ~\cite{10.1007/978-3-031-16434-7_49}.

\subsection{GFE-Net~\cite{li2023generic}}

The model was trained for 200 epochs using the Adam optimizer, with an initial learning rate of $1\times 10^{-3}$, which was gradually reduced to 0 over the final 50 epochs. The training batch size was set to 4, and all input images were resized to $256 \times 256$. Data augmentation strategies, including random horizontal and vertical flips with a probability of 0.5, were applied to enhance generalization.

We employed the same weight (e.g., $\lambda_{all}$ = 1) for all loss losses, including enhancement loss, cycle-consistency loss, and reconstruction loss. Furthermore, we adopted the architecture proposed in~\cite{li2023generic}, implementing a symmetric U-Net with 8 down-sampling and 8 up-sampling layers.

\subsection{I-SECRET~\cite{i-secret}}

The model was trained for 200 epochs using Adam optimizer with an initial learning rate of $1 \times 10^{-4}$ and $\beta$ values set to $0.5$ and $0.999$, respectively. The learning rate followed a cosine decay schedule. The training batch size was set to 8. All images were resized to $256 \times 256$ with random cropping and flipping augmentation strategies.

For model architectures, the generator consisted of 2 down-sampling layers, each with 64 filters and 9 residual blocks. Input and output channels were set to 3 for RGB inputs. The discriminator included 64 filters and 3 layers. Instance normalization and reflective padding were used. The training process employed a least-squares GAN loss~\cite{mao2017least}, a ResNet-based generator, and a PatchGAN-based~\cite{isola2017image} discriminator. GAN and reconstruction losses were weighted at $1.0$, while their importance with the contrastive loss (ICC-loss) and importance-guided supervised loss (IS-loss)~\cite{i-secret} were enabled with weights of $1.0$.

\subsection{ RFormer~\cite{deng2022rformer}.}
The model was trained for 150 epochs using Adam optimizer, with an initial learning rate of $1 \times 10^{-4}$ and $\beta$ values set to 0.9 and 0.999, respectively. The cosine annealing strategy was employed to steadily decrease the learning rate from the initial value to $1 \times 10^{-6}$ during the training procedure. The training batch size was set to 32. All images were resized to $256 \times 256$ without any additional augmentation strategies. The model architecture followed the design proposed in~\cite{deng2022rformer}, which was consistently maintained throughout our experiments.


\subsection{CycleGAN~\cite{cyclegan}, WGAN~\cite{gulrajani2017improved}, OTTGAN~\cite{wang2022optimal}, OTEGAN~\cite{zhu2023optimal}  }
The models were trained for 200 epochs using the RMSprop optimizer, with initial learning rates for the generator and discriminator set to $0.5 \times 10^{-4}$ and $1 \times 10^{-4}$, respectively. The learning rate followed a linear decay schedule, decreasing by a factor of 10 every 100 epochs. The training batch size was set to 2. All input images were resized to $256 \times 256$, with random horizontal and vertical flips applied as augmentation strategies.
For CycleGAN, the weighting parameters in the final objective were set to $\lambda_{GAN} = 1$, $\lambda_{Cycle} = 10$, and $\lambda_{Idt} = 5$, corresponding to the weights for the GAN loss, cycle consistency loss, and identity loss, respectively. The Mean Squared Error (MSE) loss was used for the GAN loss, while the cycle consistency and identity losses were computed using the L1-norm. For OTTGAN and OTEGAN, the weighting parameter $\lambda_{OT}$ was set to 40, representing the optimal transport (OT) cost. Furthermore, the OT loss was calculated using the MSE loss for OTTGAN and the MS-SSIM loss for OTEGAN. The generator and discriminator architectures were implemented following the baseline designs described in~\cite{zhu2023optimal,zhu2023otre}.

\subsection{Context-aware OT~\cite{vasa2024context}}
The model was trained for 200 epochs using the RMSprop optimizer, with initial learning rates for the generator and discriminator set to $0.5 \times 10^{-4}$ and $1 \times 10^{-4}$, respectively. The learning rate followed a linear decay schedule, decreasing by a factor of 10 every 50 epochs. The training batch size was set to 2. All input images were resized to $256 \times 256$ without additional augmentation strategies.
A warm-up training strategy was employed, wherein the context-OT loss was introduced after the first 50 epochs. The weighting parameter for this loss was set to $5\times10^{-2}$. We utilized a pre-trained VGG~\cite{mechrez2018contextual} network outlined in~\cite{vasa2024context} to compute the OT loss at feature spaces.
The generator and discriminator architectures followed the designs outlined in~\cite{vasa2024context}.

\subsection{TPOT~\cite{dong2024tpot}}
The model was trained for 100 epochs using the RMSprop optimizer over two training phases. In each phase, the learning rate was initialized at $2 \times 10^{-4}$ and reduced by a factor of 10 after
every $50$ epochs. The training batch size was set to 4, and all input images were resized to $256 \times 256$ without additional augmentation strategies. The weighting parameter for the topology regularization was fixed at 1. where the segmentation masks During training, the segmentation masks were extracted using the method proposed in~\cite{zhou2021study}. The generator and discriminator architectures followed the designs introduced in~\cite{dong2024tpot}.

\subsection{CUNSB-RFIE~\cite{dong2024cunsb}}
The model was trained for 130 epochs using the Adam optimizer, with an initial learning rate of $2 \times 10^ {-4}$. The learning rate was linearly decayed to 0 after the first 80 epochs, and the batch size was set to 8. All input images were resized to $256 \times 256$ without applying any additional augmentation strategies.

The weighting parameters in the final objective were set as $\lambda_{SB} = 1$, $\lambda_{SSIM} = 0.8$, and $\lambda_{NCE} = 1$, corresponding to the weights for entropy-regularized OT loss, task-specific regularization with MS-SSIM~\cite{brunet2011mathematical}, and PatchNCE~\cite{park2020contrastive} loss, respectively.

The generator and discriminator architectures followed the designs described in~\cite{dong2024cunsb}. Specifically, the base number of channels for the generator was set to 32, and 9 ResNet blocks were used in the bottleneck. In addition to the output features of all downsampling layers, the bottleneck's input and middle feature maps were utilized to calculate the PatchNCE regularization.

\subsection{Vessel Segmentation}
A vanilla U-Net model~\cite{ronneberger2015unet} was employed for the downstream vessel segmentation task. The network comprised 4 layers with a base channel size 64 and a channel scale expansion ratio of 2. The training was conducted over 10 epochs using the Adam optimizer, with a batch size of 64 and an initial learning rate of $5 \times 10^{-5}$, which followed a cosine annealing learning rate scheduler. 

Before training, the enhanced images and their corresponding ground-truth vessel segmentation masks were preprocessed. The preprocessing pipeline included random cropping to $ 48 \times 48$ patches, followed by data augmentation techniques such as random horizontal flips, random vertical flips (with a probability of 0.5), and random rotation.

\begin{figure}[ht]
    \centering
    \includegraphics[width=1\linewidth]{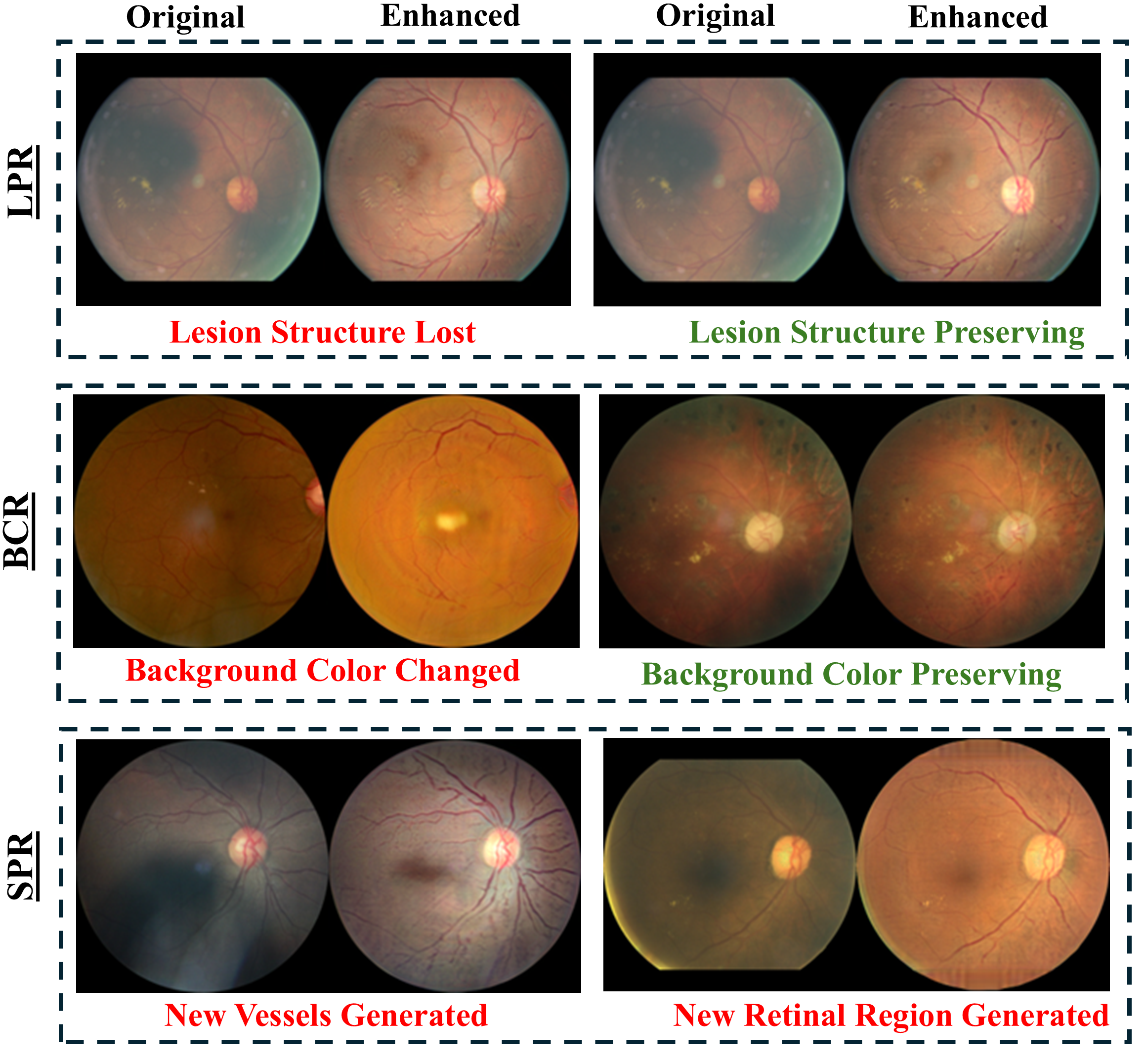}
    \caption{An illustrative medical expert clinical preference evaluation between (a) lesion preserving, (b) background preserving, and (c) structure-preserving.}
    \label{fig:expert-protocol}
\end{figure}

\begin{figure*}[t]
  \centering
  \includegraphics[width=\textwidth]{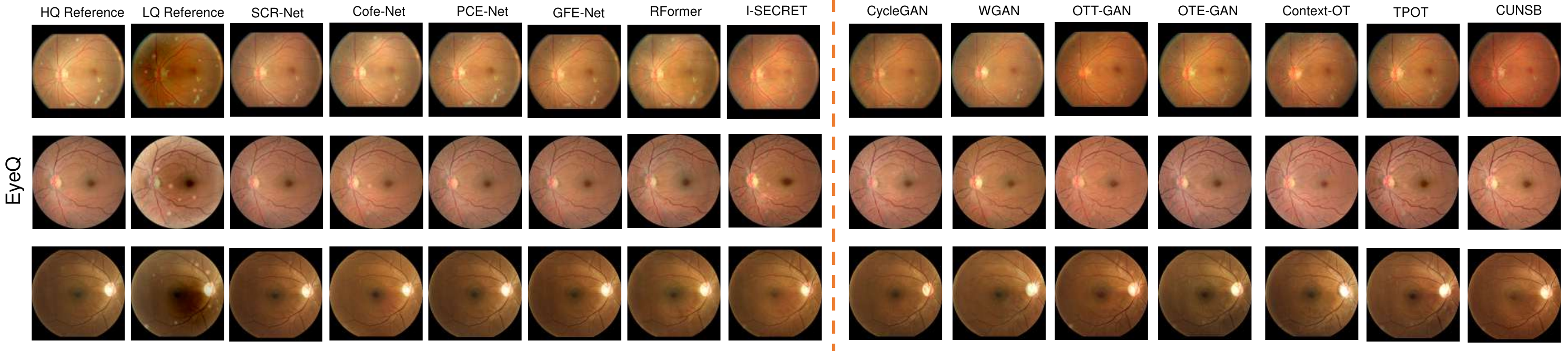}  
  \caption{ Illustration of the Denoising Evaluation on the EyeQ dataset. The first and second columns show the high- and low-quality image references, respectively, while the remaining columns display the synthetic high-quality images generated by all baseline models.
  }
  \label{fig:full-reference-eyeq}
\end{figure*}
\begin{figure*}[t]
  \centering
  \includegraphics[width=\textwidth]{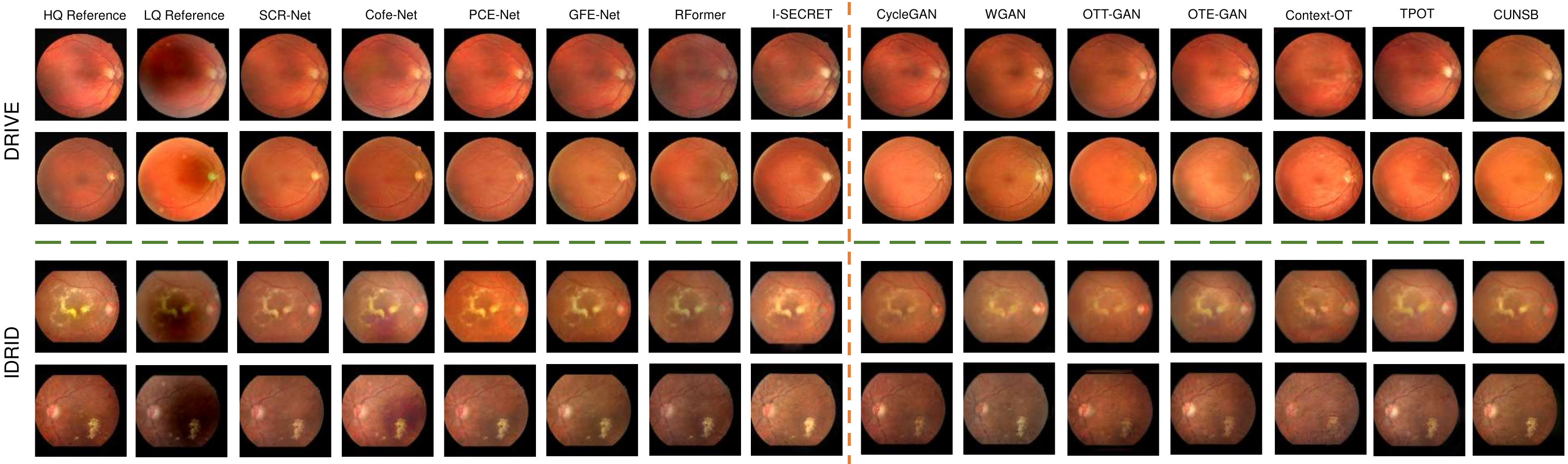}  
  \caption{ Illustration of the Denoising Generalization Evaluation on the DRIVE and IDRID datasets. The first and second columns show the high- and low-quality image references, respectively, while the remaining columns display the synthetic high-quality images generated by all baseline models.
  }
  \label{fig:full-reference-generalization}
\end{figure*}
\begin{figure*}[t]
  \centering
  \includegraphics[width=\textwidth]{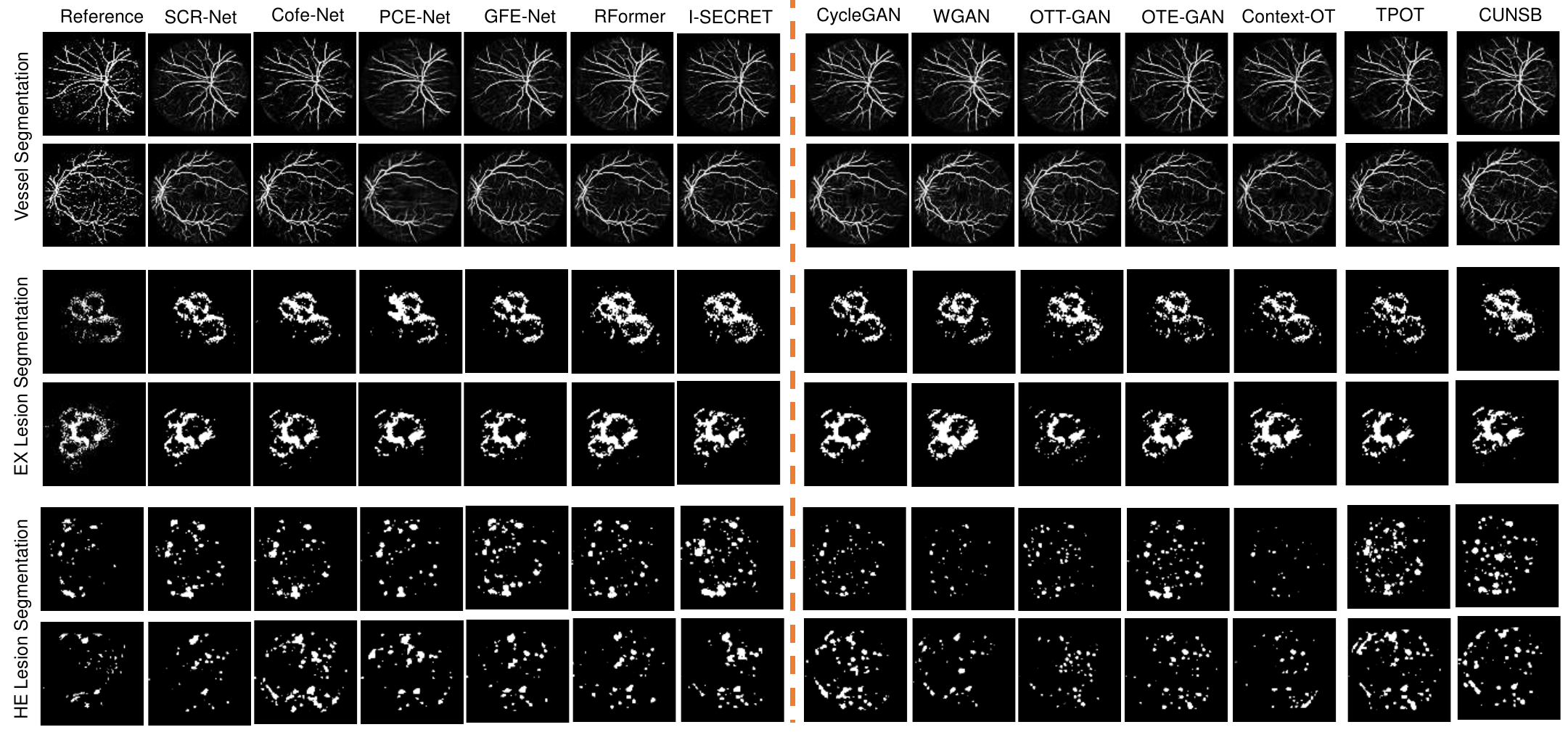}  
  \caption{ Illustration of Vessel and Lesion (EX and HE) Segmentation Experiments. The first column shows the reference segmentation masks, while the remaining columns display the segmentation results produced by all baseline models.
  }
  \label{fig:full-reference-segmentation}
\end{figure*}
\begin{figure*}[t]
  \centering
  \includegraphics[width=\textwidth]{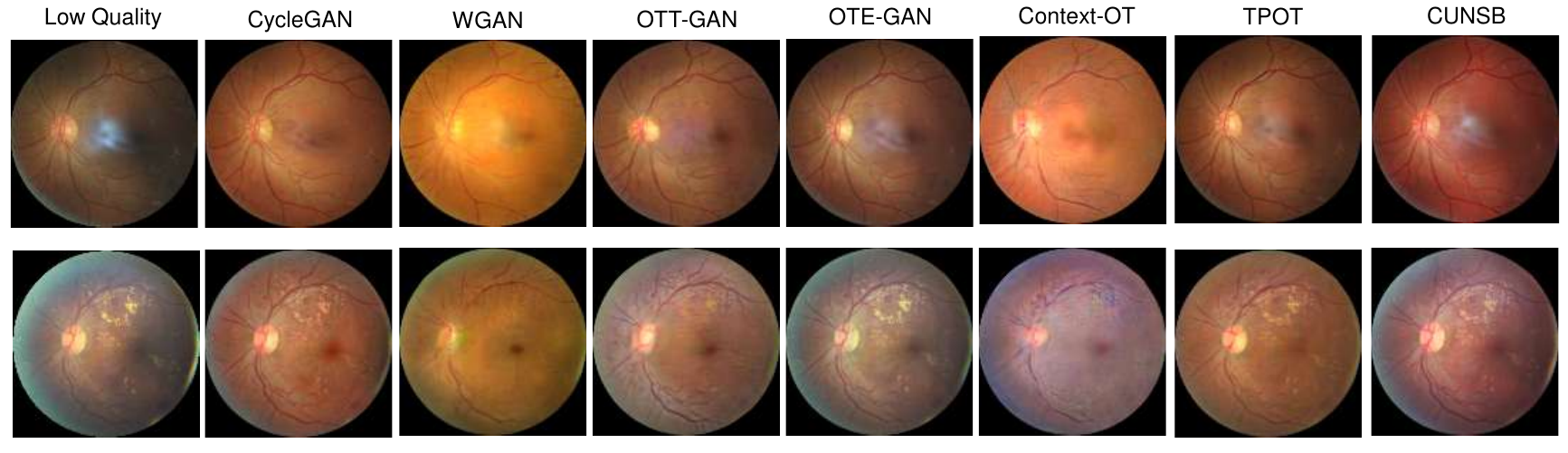}  
  \caption{ Illustration of the denoising results in the No-Reference Quality Assessment Experiments. The first column shows the input low-quality image, while the remaining columns display the synthetic high-quality images generated by all unpaired baseline models.
  }
  \label{fig:no-reference}
\end{figure*}

\section{No-Reference Quality Assessment Experiments Details}
We utilized the No-Reference Evaluation Dataset, including all unpaired baseline models for the No-Reference Assessment. These experiments evaluated the models' ability to learn and eliminate real-world noise. We maintained the experimental settings (e.g., hyperparameters) as outlined in Sec.~\ref{Sec:full-reference} to ensure a fair comparison.

\subsection{Lesion Segmentation}
Another U-Net model was employed for the downstream lesion segmentation task. The network consisted of 4 layers, with a base channel size of 64. The channel multiplier was set to 1 in the final layer and 2 in the remaining layers. The model was trained for 300 epochs using the Adam optimizer, with a batch size of 8. The initial learning rate was set to $2 \times 10^{-4}$, and a cosine annealing scheduler was applied, gradually reducing the learning rate to a minimum value of $1 \times 10^{-6}$. 

We utilized extensive data augmentation strategies to enhance model robustness. These included random horizontal and vertical flips, each with a probability of 0.5; random rotations with a probability of 0.8; random grid shuffling over $8\times 8$ grids with a probability of 0.5; and CoarseDropout, which masked up to 12 patches of size 
$20 \times 20$ to a value of 0, also with a probability of 0.5.

\subsection{DR grading.} We trained an NN-MobileNet model~\cite{deeplearning1} for the DR grading task using real-world high-quality images. The enhanced test images are used with the trained NN-MobileNet to infer DR grading classification. Enhancement performance is evaluated based on classification accuracy (ACC), kappa score, F1 score, and AUC. This evaluation primarily aims to assess whether the denoising model disrupts lesion distribution, potentially leading to inconsistencies with the original DR grading labels. 
During the training, we conducted 200 epochs with a batch size of 32 and an input size of $256 \times 256$. The AdamP optimizer was utilized with a $1 \times 10^{-3}$ weight decay and an initial learning rate of $1 \times 10^{-3}$. A dropout rate of $0.2$ was applied during training to mitigate over-fitting. Furthermore, the learning rate was dynamically adjusted using the Cosine Learning Rate Scheduler.

\subsection{Representation Feature Evaluation.}We employed two foundation models for fundus images, RetFound~\cite{zhou2023foundation} and Ret-CLIP~\cite{du2024ret}, to calculate the Fréchet Inception Distance (FID) between enhanced and real-world high-quality image feature representations. These metrics are referred to as \textit{FID-RetFound} and \textit{FID-CLIP}, respectively.

\textit{FID-Retfound}, based on a MAE backbone, captures high-level semantic structures, while \textit{FID-Clip}, trained via contrastive learning, emphasizes spatial coherence and structural consistency. To compute these metrics, the enhanced and real-world high-quality images were resized to $224 \times 224$ and normalized before being passed into the respective image encoders. The FID scores were then calculated based on the 1024-dimensional and 512-dimensional feature maps produced by RetFound and Ret-Clip, respectively.

\subsection{Experts Annotation Evaluation.} To evaluate the quality of the enhanced images, we recruited six trained specialists to conduct manual assessments. The evaluation criteria, as illustrated in Fig.~\ref{fig:expert-protocol}, included lesion preservation, background preservation, and structure preservation. Each image was individually reviewed, and the results were meticulously recorded. To minimize subjective bias, the six annotators performed cross-evaluations on test images enhanced by different models. The final annotations were further validated by ophthalmologists to ensure accuracy and clinical relevance. Despite these efforts, a degree of variability may still persist due to the inherent subjectivity and manual nature of expert evaluations.

\section{Result Illustrations}
We provide additional visualizations for all baseline models in the Full-Reference and No-Reference Quality Assessment Experiments. Specifically, Fig.~\ref{fig:full-reference-eyeq} presents the results of the Denoising Evaluation conducted on the EyeQ dataset. In contrast, Fig.~\ref{fig:full-reference-generalization} illustrates the Denoising Generalization Evaluation results on the DRIVE~\cite{drive} and IDRID~\cite{idrid} datasets. Fig.~\ref{fig:full-reference-segmentation} displays the outcomes of Vessel and Lesion (EX and HE) Segmentation. The results of the No-Reference quality assessment Experiments are outlined in Fig.~\ref{fig:no-reference}.
\end{document}